\documentclass[a4paper]{article}

\usepackage{INTERSPEECH2018}
\usepackage{cite}
\usepackage{graphicx}
\usepackage{multirow}
\usepackage{color}
\usepackage[normalem]{ulem}
\useunder{\uline}{\ul}{}

\title{Automatic Speech Recognition for Humanitarian Applications in Somali}
\name{Raghav Menon$^1$, Astik Biswas$^1$, Armin Saeb$^1$, John Quinn$^2$ and Thomas Niesler$^1$}
\address{
  $^1$Department of Electrical and Electronic Engineering, Stellenbosch University, South Africa\\
  $^2$UN Global Pulse, Kampala, Uganda}
\email{rmenon@sun.ac.za, abiswas@sun.ac.za, arsaeb@sun.ac.za, john.quinn@unglobalpulse.org, trn@sun.ac.za}

\begin{document}

\maketitle
\begin{abstract}
We present our first efforts in building an automatic speech recognition system for Somali, an under-resourced language, using 1.57 hrs of annotated speech for acoustic model training. 
The system is part of an ongoing effort by the United Nations (UN) to implement keyword spotting systems supporting humanitarian relief programmes in parts of Africa where languages are severely under-resourced. 
We evaluate several types of acoustic model, including recent neural architectures.
Language model data augmentation using a combination of recurrent neural networks (RNN) and long short-term memory neural networks (LSTMs) as well as the perturbation of acoustic data are also considered.
We find that both types of data augmentation are beneficial to performance, with our best system using a combination of convolutional neural networks (CNNs), time-delay neural networks (TDNNs) and bi-directional long short term memory (BLSTMs) to achieve a word error rate of 53.75\%. 

\end{abstract}

\section{Introduction}

Recent research has established that, in societies with accessible internet infrastructure, social media is popular for voicing views and concerns about social issues~\cite{Vosoughi_ICDMW15, Wegrzyn_CASoN11, Burnap15}. 
However, in a series of surveys by the United Nations (UN), it was found that radio phone-in talk shows become the medium of choice for such communication when internet connectivity is insufficient or inaccessible~\cite{unpulse1, unpulse2, unpulse3}.
For this reason, the development of radio browsing systems has been piloted with the UN.
Successful systems are in current active use in Uganda and monitors such radio shows in three native languages to obtain information that can inform and support relief and developmental programmes ~\cite{Menon2017, Saeb2017}. 

The success of the Ugandan system have led to calls for a similar system in Somalia. 
The acoustic models of the Ugandan system were however developed using orthographically annotated speech corpora of between 6 and 9 hours in length.
However, the urgency with which the Somali system was required, as well as the difficulty with which the required linguistic expertise could be found, allowed only for the compilation of a 1.57-hour annotated speech corpus.
This paper describes our efforts to produce the best possible Somali ASR system using this limited resource, with the intent of incorporating it in a radio browsing system for the monitoring of UN humanitarian operations in Somalia.
Due to the very small corpus, we attempt to capitalise as much as possible on available data from other languages, such as those compiled for the Ugandan system, by harnessing multilingual acoustic model training strategies.
To achieve this, we consider recent architectures, including feedforward deep neural networks (DNNs) and time-delay neural networks (TDNNs).

We have been able to locate just one account of ASR for Somali in the literature~\cite{Nimaan2006}. 
These authors trained a Somali HMM-GMM acoustic model by adapting a French acoustic model using a phoneme mapping.
A training corpus consisting of 9.5 hours of prepared Somali speech was used. In~\cite{Nimaan2006}, text-normalisation was done using tools prepared in-house to take care of the text variations. Hence, in contrast to our case, they had a well-built Somali corpus.




There has been a growing interest recently in developing ASR for under-resourced languages~\cite{Besacier2014}. 
In this paper, we follow the same direction as in~\cite{Menon2017, Saeb2017} starting out by modelling ASR with HMM-GMM and moving on to more advanced neural net based models and investigate ways in which performance in terms of word error rate (WER) can be improved for an under-resourced target language.

The paper is organized as follows: section~\ref{sec:RBS} gives a brief description about the radio browsing system, section~\ref{sec:data} provides a detailed account of the data available and language modelling followed in this paper, section~\ref{sec:AM} describes the various architectures for acoustic models along with the features used, section~\ref{sec:Exp} gives an idea of our experimental setup, section~\ref{sec:res} discusses the results obtained with section~\ref{sec:Con} concluding the paper.
 
\section{Radio Browsing System}
\label{sec:RBS}
\begin{figure}[t]
  \centering
  \includegraphics[width=\linewidth]{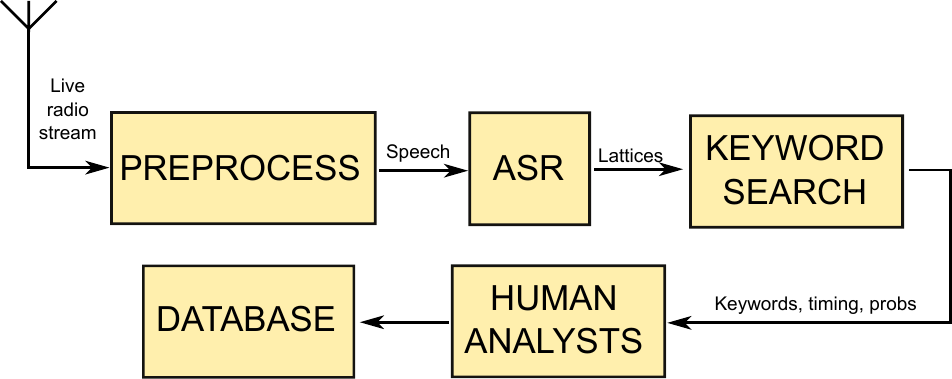}
  \caption{\it The radio browsing system~\cite{Saeb2017}.}
  \label{FIG_radio_browsing_system}
\end{figure}

The structure of the existing radio browsing system is shown in Figure~\ref{FIG_radio_browsing_system}~\cite{Saeb2017}. 
In essence, it consists of an ASR and a keyword spotter. 
The live audio stream is first pre-processed and then passed on to an ASR system where lattices are generated. 
These lattices are subsequently searched for the occurrence of the keyword under consideration. 
This output is then passed to human analysts who further processes and format the data for humanitarian decision making and situational awareness. 
This system is currently deployed and in constant use by the UN in Uganda.


\section{Data}
\label{sec:data}
The acoustic training and test data used in our experiments, as described in Table~\ref{Table1}, consisted of the following.

\begin{itemize}
\item \textbf{Somali data}, amounting to 1.57 hrs of training data and 10 minutes of test data drawn from broadcast Somali radio phone-in programmes. 
\item \textbf{Ugandan data}, consisting of radio broadcasts recorded in Ugandan English, Luganda, Acholi and Somali and amounting to 6, 9.6 and 9.2 hrs of speech respectively. 
\item \textbf{South African broadcast news data}, amounting to 20 hrs of speech recorded from a popular broadcast news channel~\cite{Kamper2015}.
\item \textbf{South African code-switched data}, consisting of 14.3 hrs of spontaneous speech obtained from broadcast soap opera episodes~\cite{van2018city}. This multilingual data contains five languages and frequent code-switching between the pairs English-isiZulu, English-isiXhosa, English-Setswana and English-Sesotho.
\end{itemize}

\begin{table}[h]
\centering
\caption{Duration in hours (h) and minutes (m) of data available for acoustic modelling. The total English data is obtained by mixing South African English and Ugandan English.}
\vspace{-6pt}
\label{Table1}
\renewcommand{\arraystretch}{0.9}
\begin{tabular}{|c|c|c|}
\hline
\textit{Set} & \textit{No of Utterances} & \textit{Duration} \\ \hline \hline
\multicolumn{3}{|c|}{\textbf{Somali (som)}} \\ \hline
Train & 1.3k & 1.57h \\ 
Test & 95 & 10m \\ \hline
Total & 1.4k & 1.74h \\ \hline \hline
\multicolumn{3}{|c|}{\textbf{English (eng)}} \\ \hline
Train & 14.9k & 26h \\ \hline \hline
\multicolumn{3}{|c|}{\textbf{Luganda (lu)}} \\ \hline
Train & 8.8k & 9.6h \\ \hline \hline
\multicolumn{3}{|c|}{\textbf{Acholi (ac)}} \\ \hline
Train & 4.9k & 9.2h \\ \hline \hline
\multicolumn{3}{|c|}{\textbf{English-isiZulu (ez)}} \\ \hline
Train & 8.4k & 4.81h \\ \hline \hline
\multicolumn{3}{|c|}{\textbf{English-isiXhosa (ex)}} \\ \hline
Train & 7k & 2.68h \\ \hline \hline
\multicolumn{3}{|c|}{\textbf{English-Setswana (et)}} \\ \hline
Train & 5.3k & 2.33h \\ \hline \hline
\multicolumn{3}{|c|}{\textbf{English-Sesotho (es)}} \\ \hline
Train & 5.7k & 2.36h \\ \hline \hline
\textbf{Total Train} & \textbf{56.3k} & \textbf{58.55h} \\ \hline
\end{tabular}%
\end{table}

All four datasets described above have been orthographically transcribed by native speakers, and pronunciation dictionaries were available.

The text corpus available for language modelling was limited is summarised in Table~\ref{Table2}. 
We compiled this corpus by harvesting 59.23k sentences from Somalian news websites, 54.85k sentences from public facebook posts (which are carefully cleaned and filtered), and 215.30k sentences from selected facebook comments (posts which were not cleaned throughly).
Furthermore, we have attempted to augment the language modelling data by artificially generating 775.3k sentences (11.29M words) by training a RNN-LSTM using the 1320 sentences in the Somali acoustic training set.
The incorporation of such artificial data has successfully improved speech recognition performance for other authors~\cite{emre2018IS}.


The language model for Somali was generated using the SRILM toolkit~\cite{Stolcke2002}. 
When training only on the language model training corpus, a perplexity of 532.71 was achieved on the transcriptions of the acoustic test set, using Kneser-Ney discounting.
We found that this discounting method was not suitable when incorporating the artificially-generated data, however.
In this case, Witten-Bell discounting was chosen, leading to a perplexity of 269.80. Language models from different sources were interpolated to obtain the final language model and the perplexities are given in Table~\ref{table:ppl}.

\begin{table}[t]
\centering
\caption{Data available for language modelling.}
\label{Table2}
\resizebox{\columnwidth}{!}{%
\begin{tabular}{|l|c|c|c|}
\hline
\multicolumn{1}{|c|}{\textbf{Corpus}} & \textbf{\begin{tabular}[c]{@{}c@{}}No of \\ Word Tokens\end{tabular}} & \textbf{\begin{tabular}[c]{@{}c@{}}No of \\ Word Types\end{tabular}} & \textbf{\begin{tabular}[c]{@{}c@{}}No of \\ Sentences\end{tabular}} \\ \hline \hline
Somali (train) & 15.1k & 4.7k & 1.3k \\
Somali (cleaned web harvested text) & 1.92M & 82.8k & 59.2k \\
Facebook post (cleaned) & 1.55M & 92.9k & 54.9k \\
Facebook comments  (uncleaned) & 3.5M & 356.7k & 215.3k \\
RNN-LSTM generated text & 11.29M & 4.7k & 775.3k \\ \hline \hline
Total & 6.98M & 433.9k & 1.1M \\ \hline
\end{tabular}%
}
\end{table}

\begin{table}[h]
\centering
\caption{Perplexity analysis on Somali test set.}
\label{table:ppl}
\resizebox{\columnwidth}{!}{%
\begin{tabular}{|l|c|c|}
\hline
\multicolumn{1}{|c|}{\textbf{Corpus}} & \textbf{\begin{tabular}[c]{@{}c@{}}Kneser-Ney\\ discounting\end{tabular}} & \textbf{\begin{tabular}[c]{@{}c@{}}Witten-Bell\\ discounting\end{tabular}} \\ \hline \hline
Somali (train) & 811.03 & 880.64 \\ \hline
Somali (cleaned web harvested text) & 502.11 & 541.86 \\ \hline
Facebook post (cleaned) & 536.43 & 571.16 \\ \hline
Facebook comments  (uncleaned) & 1062.94 & 1161.36 \\ \hline
RNN-LSTM generated text & - & 1155.51 \\ \hline
Interpolated (row 2, 3, 4, 5) & \textbf{532.71} & 555.90 \\ \hline
\begin{tabular}[c]{@{}l@{}}Interpolated (row 2, 3, 4, 5)\\ +Generated text\end{tabular} & - & \textbf{269.80} \\ \hline
\end{tabular}%
}
\vspace{-10pt}
\end{table}

\section{Acoustic Modelling}
\label{sec:AM}
The Kaldi speech recognition toolkit has been used for all experiments~\cite{Povey2011}. 
This decoder is based on finite state transducers (FSTs) and incorporates the language model, the pronunciation dictionary (lexicon) and context dependency into a single decoding graph~\cite{mohri97}. 
We have built acoustic models using hidden Markov model / Gaussian mixture model (HMM/GMM), subspace Gaussian mixture model (SGMM), deep neural network (DNN), multilingual DNN (MDNN), long short-term memory neural network (LSTM),  bidirectional LSTM (BLSTM) and time delay neural network (TDNN) architectures. 
We describe each briefly in the following.

\subsection{Hidden Markov Model}
Hidden Markov models (HMMs) with Gaussian mixture models (GMMs) observation probability densities are trained in a step-by-step fashion~\cite{Panayotov15}. 
Our HMM/GMM system includes the application of linear discriminant analysis (LDA) and maximum likelihood linear transforms (MLLT) as well as speaker adaptive training (SAT) using feature-level maximum likelihood linear regression (fMLLR) for improved acoustic modelling.
MFCC features, including velocity and acceleration coefficients ($\Delta$ and $\Delta\Delta$), were used as features.

\subsection{Subspace Gaussian Mixture Model}
Subspace Gaussian mixture models (SGMMs) allow a large collection of GMMs to be represented compactly~\cite{Povey11}.
This is advantageous especially in later model adaptation steps, such as speaker adaptation~\cite{Motlicek13}. 
SGMMs are derived from an existing HMM/GMM system by first clustering the Gaussians to obtain a universal background model (UBM).
This UBM is refined and used to initialise the SGMM, which is then optimised by EM training. 
The decoding graph obtained for this acousic model is refined by discriminative training with Boosted Maximum Mutual Information (BMMI). 
As was also done for the HMM/GMM system, fMLLR transforms were applied and MFCC features with appended $\Delta$ and $\Delta\Delta$ features were used.

\subsection{Deep Neural Networks (DNN)}
Deep neural networks (DNNs) have over the last decade emerged as a new and successful acoustic modelling technique for automatic speech recognition~\cite{yu15}. 
We employ the DNN architecture described in~\cite{Vesel2013}.
Pre-training was not performed and the weight initialisation was random. 
Our DNN uses five 2048-dimensional hidden layers and an output dimension of 1037.
The learning rate was allowed to decrease exponentially during training.
As with the HMM/GMM and SGMM systems, fMLLR was applied. 
As features, 23-dimensional log mel-filterbank energies with appended $\Delta$, $\Delta\Delta$ and 3-dimensional pitch were used. 

\subsection{Multilingual Deep Neural Networks (MDNN)}
\begin{figure}[t]
  \centering
  \includegraphics[scale=0.088]{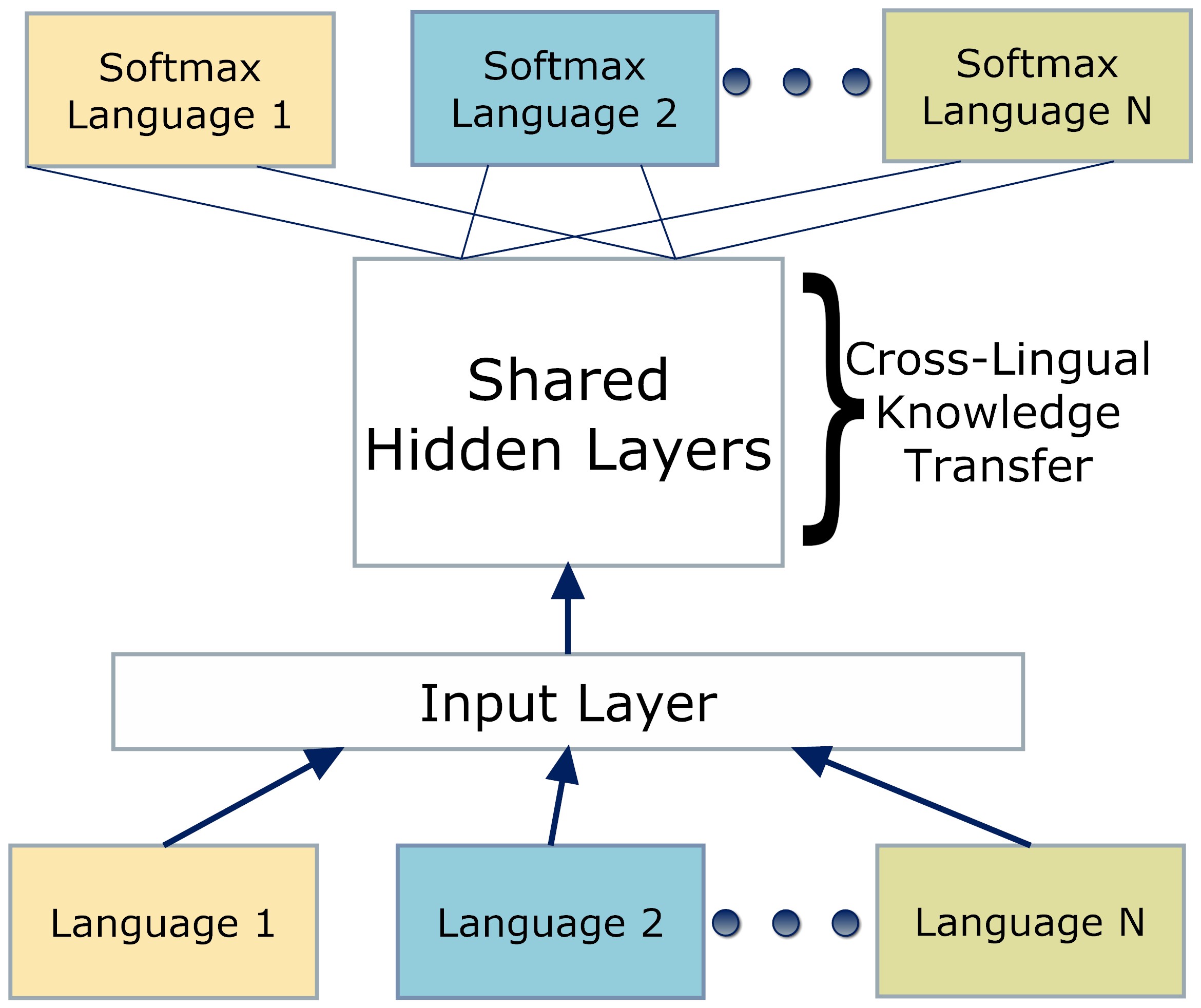}
  \caption{\it Multilingual DNN/HMM acoustic model~\cite{biswas2018IS}.}
  \label{FIG_MDNN_HMM}
\end{figure}

DNNs perform poorly as acoustic models when the training set is small, as is necessarily the case for under-resourced languages. 
This can be addressed by multilingual training~\cite{Grezl2014_1,Mohan2015,Sahraeian2016}. 
Here the DNN is trained using a large training corpus containing multiple languages to provide the class conditional posterior probabilities required by HMMs. 
Figure~\ref{FIG_MDNN_HMM} illustrates one possible multilingual architecture, in which the DNN shares hidden layers across all languages~\cite{Huang2013, Saeb2017}. 
Each language has its own softmax layer and its own HMM. 
As features, 23-dimensional log mel-filterbank energies with appended $\Delta$, $\Delta\Delta$ and 3-dimensional pitch were used.

\subsection{Long Short-Term Memory (LSTM) and Bi-directional LSTMs (BLSTM)}
\begin{figure}[t]
  \centering
  \includegraphics[width=\columnwidth]{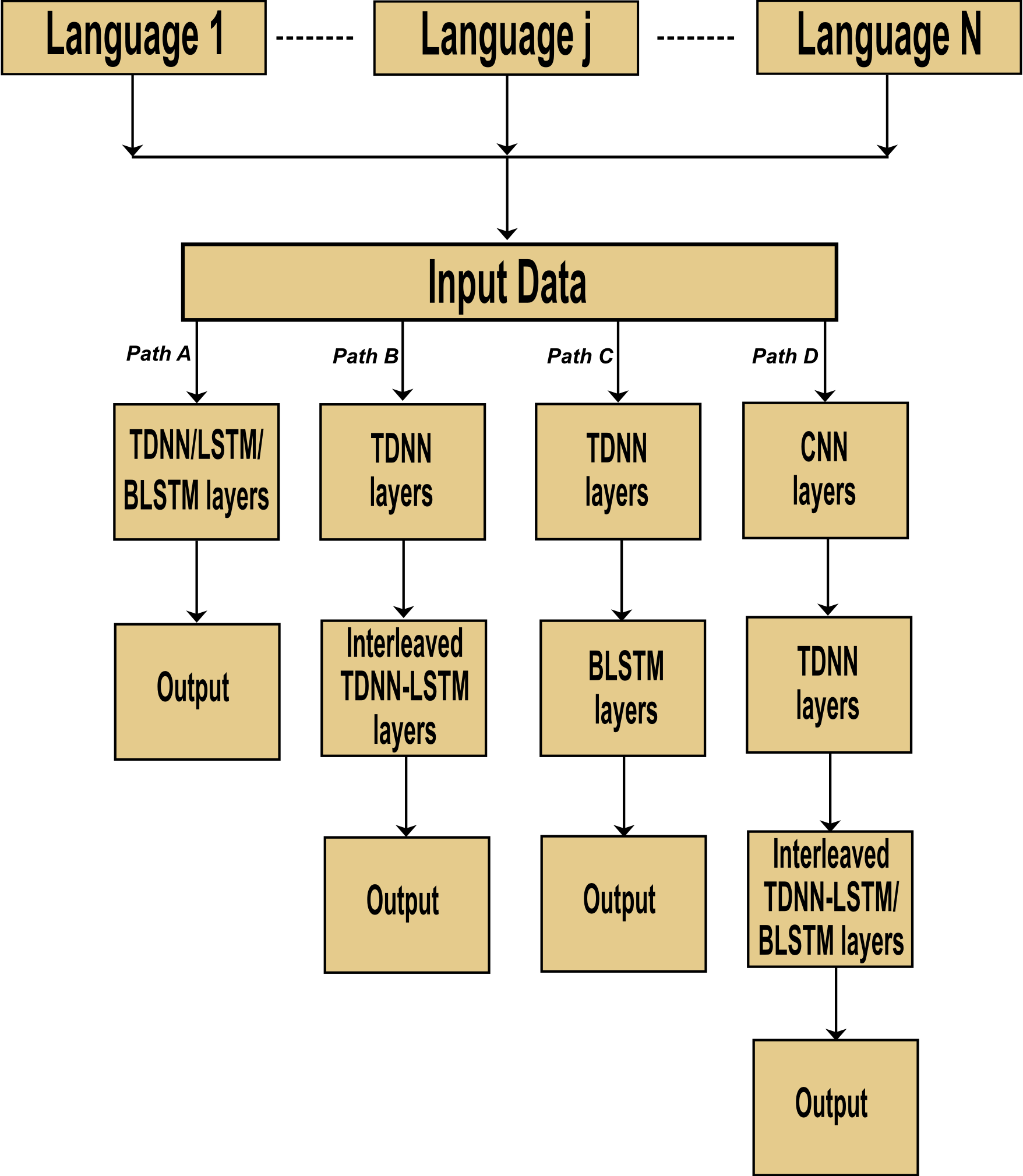}
  \caption{\it Neural network architectures considered for acoustic modelling.}
  \label{FIG_NN_Arch}
  \vspace{-10pt}
\end{figure}

Acoustic modelling with long short-term memory (LSTM) has been successfully used and shown to be efficient~\cite{SakSB14, geiger2014robust, LiW14a}. 
A LSTM uses a memory block whose behavior is controlled by an input gate, a forget gate and an output gate. 
This configuration successfully exploits long-term temporal context dependencies. 
The memory blocks are recurrently connected to form a network. 
When the information is allowed to flow both forward and backward in time, a bi-directional LSTM (BLSTM) is obtained. 
We perform multilingual training by pooling the acoustic data presented in Table~\ref{Table1} and presenting it to a LSTM/BLSTM network. The network outputs are phone targets.
The trained network is subsequently adapted to the target language. 
Speaker adaptation is achieved using i-vectors. 
A 40-dimensional MFCC with appended 3-dimensional pitch and 100-dimensional i-vector is used as feature. 
This architecture is indicated by Path A in Figure~\ref{FIG_NN_Arch}.

\subsection{Time-delay neural network (TDNN)-based models}
Time-delay neural networks (TDNN)  were introduced to overcome the limitation of DNNs in dealing with the dynamic nature of speech~\cite{waibel1989, waibel1989_1}. 
A TDNN is able to represent temporal relationships between features by introducing delays on each input and processing each of these delays with different weights. 
The TDNN architecture is modular and incremental.
Small networks are trained to perform limited tasks develop time invariant, hidden abstractions.
These are subsequently used to train larger networks~\cite{waibel1989_1}. 
The limitation of TDNN systems has been that they are computationally demanding to the extent that even GPUs could not sufficiently accelerate training. 
This has been overcome by the introduction of a subsampling procedure that assumes that neighbouring activations are correlated in time\cite{peddinti2015time}.
Such subsampling has been demonstrated to not only make the computational complexity of TDNNs practical, but also to improve ASR performance. 

We have evaluated TDNN, TDNN/LSTM and TDNN/BLSTM network architectures for acoustic modelling, indicated respectively by Paths A, B and C in Figure~\ref{FIG_NN_Arch}.
We have also considered the incorporation of a convolutional layer at the input of the TDNN/LSTM and TDNN/BLSTM networks, and indicate this by Path D in Figure~\ref{FIG_NN_Arch}. 
Multilingual training is achieved by pooling the acoustic data presented in Table~\ref{Table1} during training and subsequently adapting the trained network to the target language. 
The features were 40-dimensional MFCCs with appended 3-dimensional pitch and 100-dimensional i-vectors for speaker adaptation.

\section{Experimental Setup}
\label{sec:Exp}
 
We have modified the Kaldi 'wsj' recipe for our TDNN, LSTM and BLSTM experiments. For CNN-TDNN-LSTM experiment we used the Kaldi 'ami' recipe.
For the CNN-TDNN-BLSTM experiment, we have adapted the Kaldi 'ami' recipe by replacing the interleaved TDNN-LSTM layer with BLSTM layers.
While in all cases the number of layers was unchanged, hyperparameters such as the learning rate range, context definitions, number of epochs and chunk widths were optimised. 
Monolingual acoustic models were built using only the Somali data. 
To obtain multilingual models, we incorporate the additional languages shown in Table~\ref{Table1}. 
Since the performance of neural networks is known to improve with the amount of training data, speed perturbation (using factors 0.9, 1.0 and 1.1) and dithering is performed before training the TDNN-based models. 
System performance is reported in terms of word error rate (WER).

\section{Results and Discussion}
\label{sec:res}

Table~\ref{Table3} indicates WERs achieved using various acoustic models. 
Decoding was performed using language models built with and without the artificially generated text. 
From the table we see that, for HMM/GMM and SGMM, acoustic models, artificially generated text seems degrade performance slightly.
In contrast, the incorporation of the artificially generated text in the language model results in small improvements for the neural networks acoustic models. 
We note that the parameters used for RNN-LSTM text generation were not optimised and it is possible that further improvements could be achieved if this were done.
 
\begin{table}[h]
\renewcommand{\arraystretch}{1.2}
\centering
\caption{Somali word error rates (WER) achieved using different acoustic modelling methods and different training data.}
\label{Table3}
\resizebox{\columnwidth}{!}{%
\begin{tabular}{|l|c|c|}
\hline
\multicolumn{1}{|c|}{\multirow{2}{*}{\textbf{Acoustic Model}}} & \multicolumn{2}{c|}{\textbf{WER (\%)}} \\ \cline{2-3} 
\multicolumn{1}{|c|}{} & \textit{\textbf{\begin{tabular}[c]{@{}c@{}}No RNN-LSTM- \\ generated text\end{tabular}}} & \textit{\textbf{\begin{tabular}[c]{@{}c@{}}With RNN-LSTM- \\generated text\end{tabular}}} \\ \hline \hline
\textbf{HMM/GMM} (som) & 63.71 & 63.92 \\ 
\textbf{SGMM}    (som) & 60.78 & 60.98 \\ 
\textbf{DNN}     (som) & 59.82 & 59.48 \\ 
\begin{tabular}[c]{@{}l@{}}\textbf{MDNN}  (som+eng+lu+ac)\end{tabular} & 56.75 & 56.21 \\ 
\begin{tabular}[c]{@{}l@{}}\textbf{MDNN}  (som+eng+lu+ac+ez+ex)\end{tabular}      &-& 55.93                                                   \\ 
\begin{tabular}[c]{@{}l@{}}\textbf{MDNN}  (som+eng+lu+ac+az+ex+et+es)\end{tabular} &-& 57.23                                                   \\ \hline
\end{tabular}%
}
\end{table}

We also see from Table~\ref{Table3} that the DNN/HMM architecture improved on the WER of the HMM/GMM by 4.44\% absolute. 
Furthermore, a multilingual DNN (MDNN) incorporating the additional English, Luganda and Acholi data indicated in Table~\ref{Table1} achieved a further improvement of 3.27\% over the DNN/HMM.
This indicates that the neural network was able to take advantage of the larger training set afforded by the additional and unrelated languages shown in Table~\ref{Table1}.
Subsequent adaptation of the MDNN to the Somali data did not yield further gains.

When the code-switched data is added to the training set, the results are mixed.
A small improvement in WER (0.28\%) is achieved when adding the English-isiZulu and English-isiXhosa data.
However, the further addition of English-Setswana and English-Sesotho is seen to degrade the WER by 1.02\%. 
This indicates that adding to the pool of languages is not always beneficial, and that care must be taken when deciding on the composition of the multilingual training set.

Table~\ref{Table5} shows the speech recognition performance achieved using LSTM, BLSTM and TDNN-based acoustic models. 
In all cases the training data consisted of the English, Luganda, Acholi and Somali listed in Table~\ref{Table1}.
All models were subjected to adaptation using the Somali data after training, and in contrast to the MDNN models, this achieved absolute improvements of between 1 and 5\%. 
The LSTM model was outperformed by the TDNN model both with and without adaptation.
The BLSTM fared even worse than the LSTM without adaptation, but better than both the LSTM and TDNN after adaptation.
The TDNN-LSTM and TDNN-BLSTM both fared better than the LSTM and BLSTM models respectively, with and without adaptation.
The addition of the CNN layer led to further consistent improvements for both TDNN-LSTM and TDNN-BLSTM architectures.
Our best system used a CNN-TDNN-BLSTM acoustic model and achieved a word error rate of 53.75\%.
Experiments incorporating the code-switched languages were unfortunately still in progress at the time of paper submission.

\begin{table}[!h]
\renewcommand{\arraystretch}{1.2}
\centering
\caption{Somali word error rates (WER) achieved using multilingual LSTM, BLSTM and various TDNN-based acoustic models trained on Somali, English, Luganda and Acholi.}
\label{Table5}
\begin{tabular}{|l|c|c|}
\hline
\multicolumn{1}{|c|}{\multirow{2}{*}{\textbf{Acoustic Model}}} & \multicolumn{2}{c|}{\textbf{WER (\%)}}                                                                             \\ \cline{2-3} 
\multicolumn{1}{|c|}{}                                & \multicolumn{1}{l|}{\textit{\textbf{No Adaptation}}} & \multicolumn{1}{l|}{\textit{\textbf{With Adaptation}}} \\ \hline \hline
\textbf{TDNN}                               & 60.57                                                & 58.05                                                  \\ 
\textbf{LSTM}                                & 61.80                                                & 60.23                                                  \\ 
\textbf{BLSTM}                               & 62.14                                                & 57.91                                                  \\ 
\textbf{TDNN-LSTM}                           & 59.21                                                & 58.05                                                  \\ 
\textbf{TDNN-BLSTM}                          & 58.12                                                & 56.41                                                  \\ 
\textbf{CNN-TDNN-LSTM}                       & 56.21                                                & 55.53                                                  \\ 
\textbf{CNN-TDNN-BLSTM}                      & 55.12                                                & 53.75                                                  \\ \hline
\end{tabular}
\vspace{-10pt}
\end{table}

\section{Conclusion}
\label{sec:Con}
We have presented our first efforts in building an automatic speech recognition system for Somali as part of a United Nations programme aimed at humanitarian monitoring.
A corpus of only  1.57h of in-domain Somali data was available.
In addition, approximately 59h of annotated speech in 7 unrelated languages was available for multilingual modelling.
Several acoustic model architectures were evaluated, including state-of-the-art neural configurations and a new combination of CNN, TDNN and BLSTM layers. 
Best performance was achieved by a multilingually-trained CNN-TDNN-BLSTM acoustic model, achieving a word error rate of 53.75\%. 
Augmentation of the language model training data using a generative RNN-LSTM afforded small gains.
We also found that BLSTM layers consistently benefited more from subsequent adaptation to the target language than LSTM layers.

\section{Acknowledgements}
We thank the NVIDIA corporation for the donation of GPU equipment used for this research. We also gratefully acknowledge the support of Telkom South Africa.

\bibliographystyle{IEEEtran}

\bibliography{mybib}

\end{document}